\documentclass{Interspeech}
\usepackage{graphicx}
\usepackage{subcaption}
\usepackage{amsmath}
\usepackage{cancel}
\DeclareMathOperator*{\argmax}{arg\,max}
\usepackage[
backend=biber,
style=ieee,
citestyle=numeric-comp,
maxbibnames=1,
maxcitenames=1,
doi=false,isbn=false,url=false,eprint=false
]{biblatex}

\defbibheading{bibliography}[\refname]{}

\interspeechcameraready

\title{Chain-of-Thought Training for Open E2E Spoken Dialogue Systems}

\author[affiliation={1}]{Siddhant}{Arora}
\author[affiliation={1}]{Jinchuan}{Tian}
\author[affiliation={2}]{Hayato}{Futami}
\author[affiliation={1}]{Jee-weon}{Jung$^{\dagger}$}
\author[affiliation={1}]{Jiatong}{Shi}
\author[affiliation={2}]{Yosuke}{Kashiwagi}
\author[affiliation={2}]{Emiru}{Tsunoo}
\author[affiliation={1}]{Shinji}{Watanabe}

\affiliation{Language Technologies Institute}{Carnegie Mellon University}{USA}
\affiliation{}{Sony Group Corporation}{Japan}
\email{siddhana@andrew.cmu.edu}
\usepackage{pifont}
\usepackage{tikz}
\newcommand{\xmark}{\ding{55}}%
\newcommand*\circled[1]{\tikz[baseline=(char.base)]{
            \node[shape=circle,draw,inner sep=0.4pt] (char) {#1};}}

\usepackage{comment}

\begin{document}

\maketitle

\begin{abstract} 
Unlike traditional cascaded pipelines, end-to-end (E2E) spoken dialogue systems preserve full differentiability and capture non-phonemic information, making them well-suited for modeling spoken interactions. However, existing E2E approaches often require large-scale training data and generates responses lacking semantic coherence.
We propose a simple yet effective strategy leveraging a chain-of-thought (CoT) formulation, ensuring that training on conversational data remains closely aligned with the multimodal language model (LM)'s pre-training on speech recognition~(ASR), text-to-speech synthesis (TTS), and text LM tasks.
Our method achieves over 1.5 ROUGE-1 improvement over the baseline, successfully training spoken dialogue systems on publicly available human-human conversation datasets, while being compute-efficient enough to train on just 300 hours of public human-human conversation data, such as the Switchboard.
We will publicly release our models and training code.
\end{abstract}

\renewcommand{\thefootnote}{\fnsymbol{footnote}}
\footnotetext[2]{Currently at Apple.}
\renewcommand{\thefootnote}{\arabic{footnote}}

\section{Introduction}

Spoken dialogue systems~\cite{jokinen2009spoken,breazeal2008social} are designed to engage in natural and interactive conversations with end users, playing a critical role in voice assistants and intelligent home devices. Despite their growing importance, building effective spoken dialogue systems remains a challenging task due to the complexity of human communication.
Traditionally, spoken dialogue systems~\cite{glass1999challenges,huang2024audiogpt} comprise multiple modules, including voice activity detection (VAD)~\cite{pywebrtcvad}, automatic speech recognition (ASR)~\cite{whisper,owsm}, natural language understanding (NLU)~\cite{mehri2020dialoglue}, natural language generation (NLG)~\cite{peng-etal-2020-shot}, and text-to-speech (TTS) synthesis~\cite{coqui_openxtts}. Each of these components presents unique challenges.
For instance, the ASR module must accurately process shorter, spontaneous speech with disfluencies and filler words~\cite{Switchboard}. Additionally, spoken dialogue systems must be capable of understanding~\cite{rashkin2018towards,hara2018prediction,ward2000prosodic}, and generating~\cite{sundaram2007automatic,fujie2004conversation} non-phonemic information, like emotions, to create natural interactions. 

\begin{figure*}[t]
\centering
    \begin{subfigure}{0.37\textwidth}
        \centering
        \includegraphics[width=\linewidth]{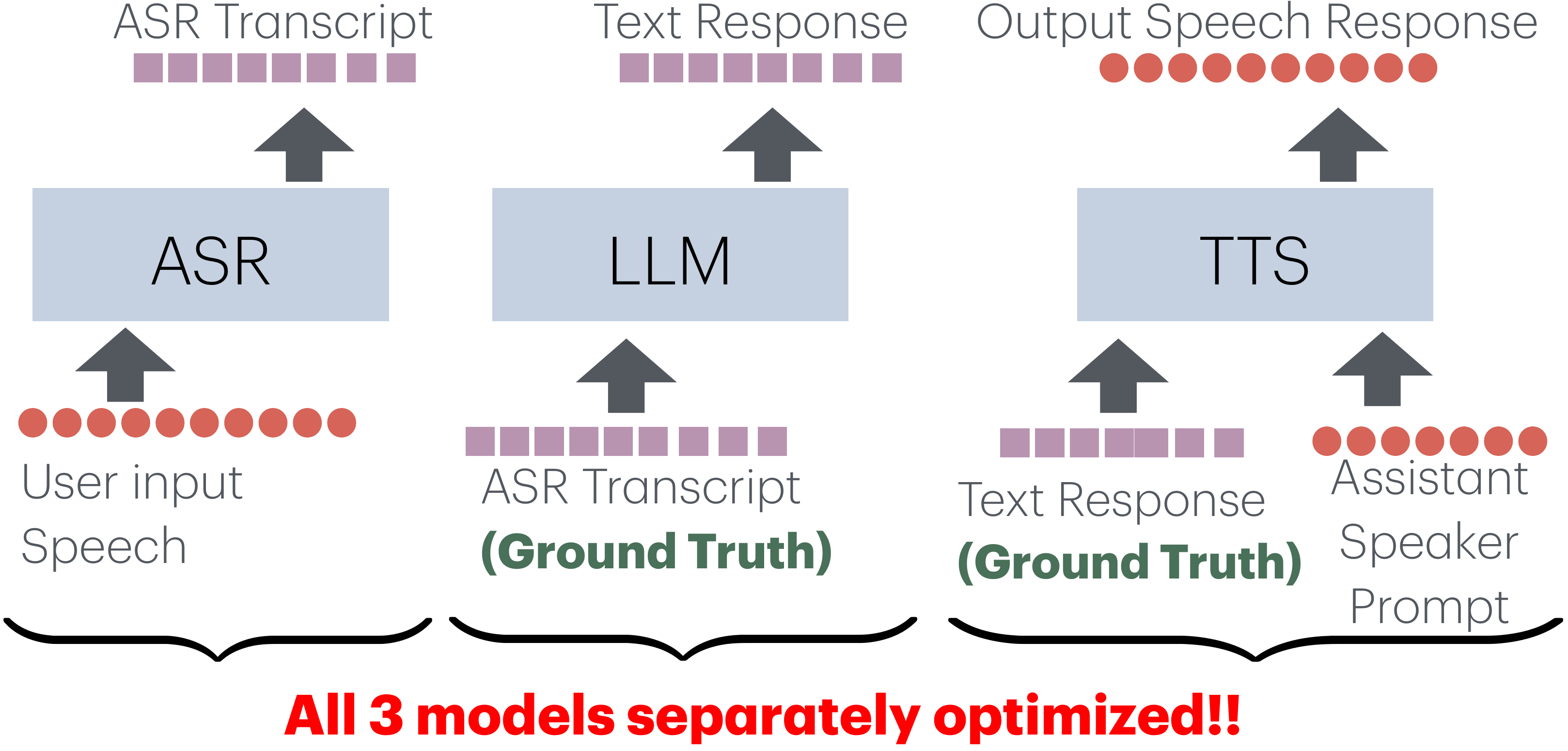} %
        \vskip -0.05in
        \caption{Cascaded Spoken Dialog System}
        \label{fig:cascaded}
    \end{subfigure}
    \begin{subfigure}{0.24\textwidth}
        \centering
        \includegraphics[width=\linewidth]{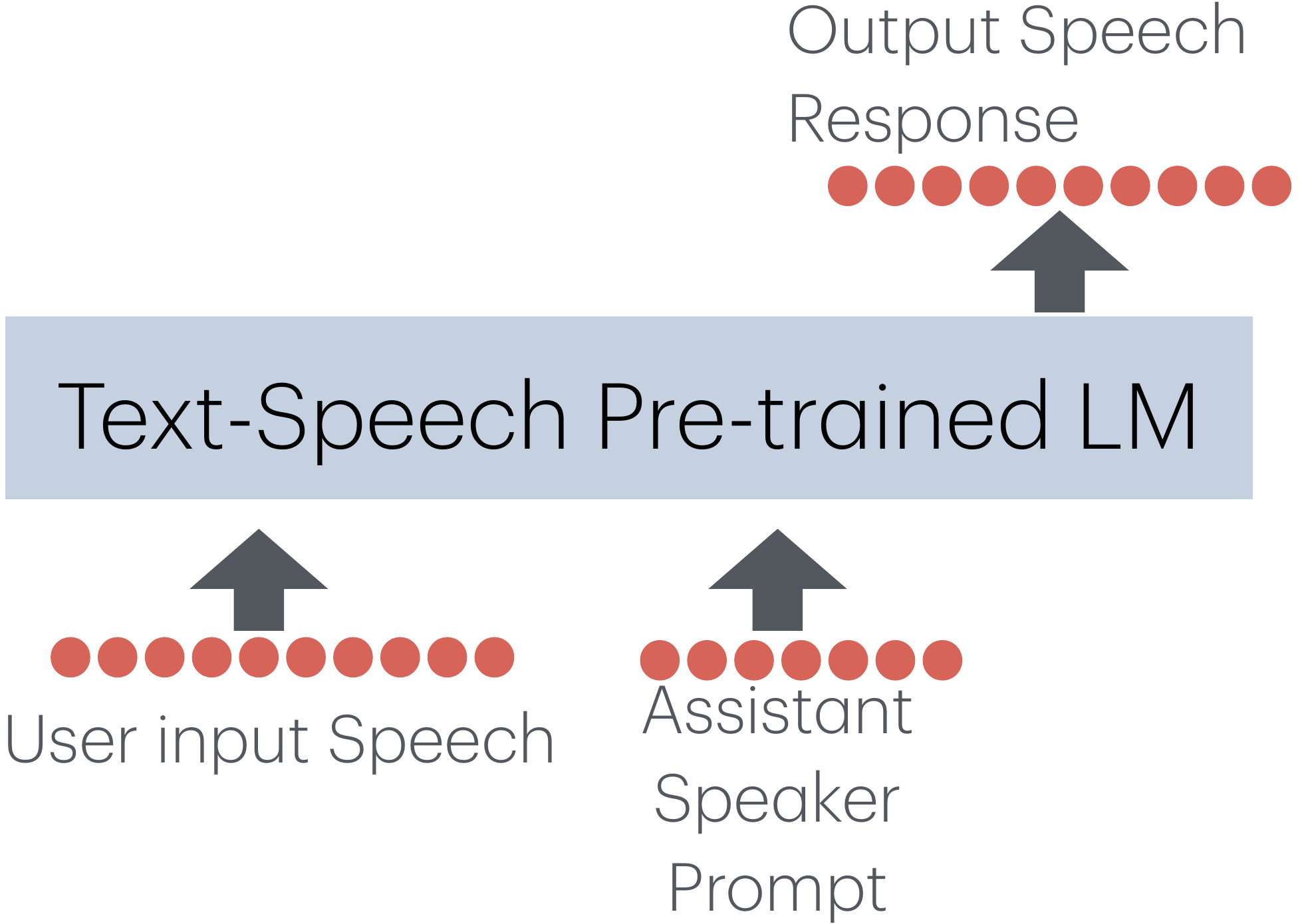} %
        \vskip -0.05in
        \caption{E2E Spoken Dialog System}
        \label{fig:e2e}
    \end{subfigure}
    \begin{subfigure}{0.378\textwidth}
        \centering
        \includegraphics[width=\linewidth]{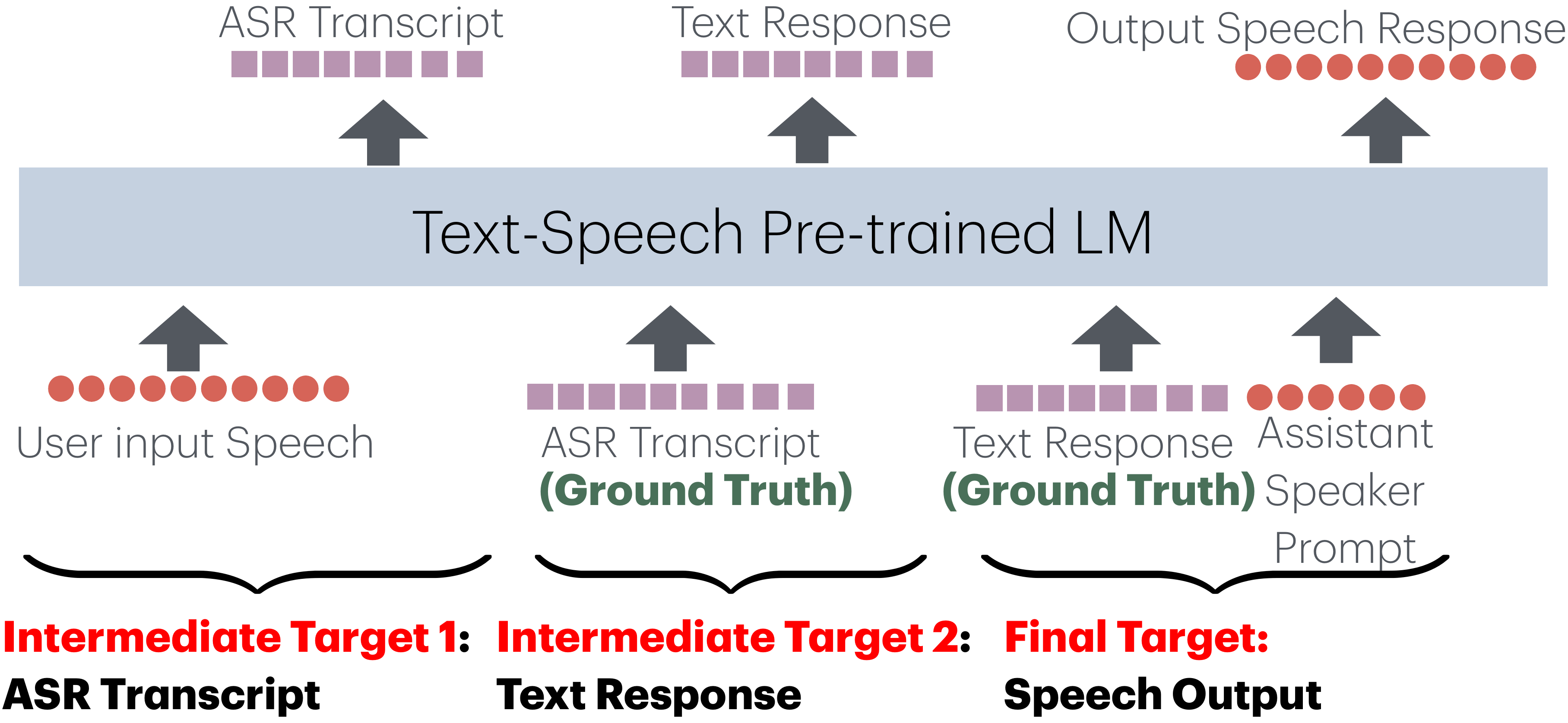} %
        \vskip -0.05in
        \caption{(Proposed) CoT E2E Spoken Dialog System}
        \label{fig:cot_e2e}
    \end{subfigure}
    \vskip -0.1in
\caption{Training schematic for our CoT-based E2E spoken dialogue system, compared with cascaded and conventional E2E systems. Our approach employs multi-stage reasoning, improving semantic coherence and speech quality while preserving E2E differentiability. We use teacher forcing during training, and multi-stage decoding using generated intermediate outputs for inference (Sec.~\ref{subsec:training_and_inference}). }
\label{fig:system-overview}
\vskip -0.23in
\end{figure*}

Recently, E2E spoken dialogue systems\footnote{\url{https://openai.com/index/hello-gpt-4o/}, \url{https://deepmind.google/technologies/gemini/}}~\cite{zhang2023speechgpt,xie2024miniomnilanguagemodelshear} have been proposed to process the user's input speech utterance and generate an appropriate speech response within a single unified architecture. This approach avoids error propagation and can better capture non-phonemic cues such as emotion.
Most existing E2E spoken dialogue systems~\cite{kyutai2024moshi} rely on complex architectures that significantly diverge from the model’s pre-training setup, requiring extensive training compute (7 million hours of unlabelled audio data for multi-stream post-training and 20K hours of speech conversation data). Additionally, the lack of structured reasoning in E2E systems can often lead to less coherent responses, which we will reveal in our experiments (Sec.~\ref{subsec:semantic_audio_result}).

In this work, we \circled{1} introduce a chain-of-thought (CoT)~\cite{wei2022chain,zhang2022automatic} training approach, restructuring post-training into a composition of ASR, text response generation, and TTS sequences. 
We apply CoT post-training to an open multimodal LLM~\cite{jinchuan2024speechlm} pre-trained on ASR, TTS, and Text LM tasks, aligning with its pre-training process to enhance compatibility and enable efficient adaptation to future architectures.
\circled{2} Our experiments show that this efficient setup builds an effective conversational system using publicly available datasets, achieving strong results with as little as 300 hours of human-human conversation dataset Switchboard~\cite{Switchboard}. The proposed post-training approach delivers faster convergence, lower compute costs, and better data efficiency, which is crucial for low-resource domains like healthcare, where large datasets are difficult to collect.
\circled{3} We compare our method with cascaded baselines, where the same pre-trained SpeechLM is fine-tuned separately for each sub-task. Results show that our CoT-based E2E model achieves comparable performance while being 3× more parameter-efficient.
\circled{4}~An ablation study confirms that CoT-based training improves both semantic coherence and synthesized speech quality compared to traditional E2E systems. Additionally, our model better preserves emotional nuances in generated speech, showing higher emotional similarity to human references.
\circled{5} Finally, we will release an open-source framework for training and evaluating cascaded and E2E dialogue systems using open source SpeechLMs on public datasets.

\section{Related studies}
Recently, ``Chat'' SpeechLMs~\cite{ji2024wavchatsurveyspokendialogue} have gained attention for their ability to engage in natural and interactive conversations. Early efforts~\cite{chu2023qwen,fu2024vita,held2024distilling} developed speech-aware LMs that generated text responses from spoken input but relied on external TTS and VAD systems for spoken dialogue.
More recently, speech-to-speech LMs~\cite{zhang2024speechgpt,Dialog_GSLM,veluri2024beyond,fang2024llama,meng2024parrot} have emerged, capable of both understanding and generating speech within a single architecture.

SpeechGPT~\cite{zhang2023speechgpt} employs a CoT instruction-tuning approach to enhance cross-modal conversational capabilities. 
However, SpeechGPT is trained solely on large-scale synthetic speech due to the difficulty of preparing speech instruction data.
In contrast, we perform CoT training on real conversational data, showcasing its practical applicability and effectiveness in low-resource scenarios.
This benefit comes from the explicit incorporation of ASR, text-to-text, and TTS task tokens~(Sec.~\ref{sec:method}), enabling a more modular and efficient CoT training process.
While OMNI-Flatten~\cite{zhang2024omniflatten} introduced a half-duplex training approach, where the model sequentially performs ASR, text response generation, and TTS. Although their setup resembles ours, their half-duplex approach serves only as an intermediate training step and lacks CoT reasoning during inference. 
In contrast, our work presents a novel investigation into the efficacy of CoT E2E spoken dialogue systems, providing a comprehensive comparison with standard cascaded and E2E methodologies using an open-source toolkit.

\section{Method}
\label{sec: problem formulation}
We begin our problem formulation with a single-turn spoken dialog system, where $X$ and $Y$ represent the user's and system's speech feature sequences.
Traditional cascaded spoken dialogue systems employ separate modules for sub-tasks, as shown in Fig.~\ref{fig:cascaded}.
These systems typically consist of an ASR model, which optimizes $P(S^{\text{asr}} | X)$ to produce the ASR transcript $\hat{S}^\text{asr}$; a text response generation model based on LLM which optimizes $P(S^{\text{res}} | \hat{S}^{\text{asr}})$ to estimate the text response $\hat{S}^\text{res}$; and a TTS model, which optimizes $P(Y | \hat{S}^{\text{res}}, X^{\text{spk}})$, where $X^{\text{spk}}$ is the TTS speaker prompt. Since each module is separately optimized, the system suffers from error propagation.
To address this, E2E spoken dialogue systems~\cite{zhang2023speechgpt,kyutai2024moshi} maximize the posterior distribution $P(Y|X,X^{\text{spk}})$ as shown in Fig.~\ref{fig:e2e}. While these systems offer fully E2E differentiable training, they lack structured reasoning, leading to less coherent responses and increased training data requirements.

\vspace{-3pt}
\subsection{CoT formulation of E2E Spoken Dialog Systems}
\vspace{-3pt}
\label{sec:method}
Inspired by the strengths of both schools of thought, our method aims to perform CoT training by explicitly incorporating the prediction of ASR transcript $S^\text{asr}$ and text response $S^\text{res}$ within the E2E spoken dialogue formulation. By regarding the $S^\text{asr}$ and $S^\text{res}$ as a probabilistic variable, we can theoretically incorporate them into the posterior distribution $P(Y|X,X^{\text{spk}})$ (\S~\ref{sec: problem formulation}) via sum rule. We further use Viterbi approximation and conditional independence (C.I.) assumption to get:
\begin{align}
\notag
P(Y | X, X^{\text{spk}}) &\approx 
P(Y | X, X^{\text{spk}}, \hat{S}^{\text{res}}, \hat{S}^{\text{asr}}) \\ \notag
&\quad \text{where
$\hat{S}^{\text{res}} = \argmax_{S^{\text{res}}} p(S^{\text{res}}|X, \cancel{X^{\text{spk}}}, \hat{S}^{\text{asr}})$} \\
&\quad \text{where
$\hat{S}^{\text{asr}} = \argmax_{S^{\text{asr}}} p(S^{\text{asr}}|X, \cancel{X^{\text{spk}})}$}
\label{eq:final_formulation}
\end{align}
To realize the formulation described in Eq.\ref{eq:final_formulation}, this work proposes a CoT E2E spoken dialog system, as shown in Figure~\ref{fig:cot_e2e}. The architecture is built around a pre-trained decoder-only LM capable of both processing and generating audio and text tokens. The \textbf{CoT model} follows a structured decoding process: \textbf{ASR Decoding}: Generates the ASR transcript $\hat{S}^{\text{asr}}$ by modeling $ P(S^{\text{asr}} | X) $. \textbf{Text Response Prediction}: 
Predicts the text response $\hat{S}^{\text{res}}$, optimizing $ P(S^{\text{res}} | X, \hat{S}^{\text{asr}}) $, akin to a cascaded system but additionally conditions on input speech 
$X$.
\textbf{Final Speech Output Generation}: Generates the speech output $\hat{Y}$ by modelling $ P(Y | X, X^{\text{spk}}, \hat{S}^{\text{asr}}, \hat{S}^{\text{res}})$.

Compared to standard E2E systems that model $P(Y | X, X^{\text{spk}})$, our CoT model preserves E2E differentiability while improving semantic coherence and speech quality through multi-stage reasoning, conditioning on intermediate outputs $\hat{S}^{\text{asr}}$ and $\hat{S}^{\text{res}}$. Compared to TTS in the cascaded system that model $P(Y | \hat{S}^{\text{res}}, X^{\text{spk}})$, our approach reduces error propagation and improves expressive speech synthesis by conditioning on input speech $X$ and ASR transcript $\hat{S}^{\text{asr}}$ to generate final speech output. We apply the proposed CoT training as a post-training strategy on an open SpeechLM~\cite{jinchuan2024speechlm} that has been pre-trained on ASR, TTS, Speech Continuation, and TextLM tasks. By aligning each reasoning stage with the SpeechLM’s pre-training tasks—namely ASR, text LM, and TTS — our approach achieves faster convergence and improved training data efficiency.
 \begin{table}[t]
 \caption{Task templates for ASR, Text Response Generation (T2T), TTS, and both E2E and CoT E2E (Proposed) spoken dialogue models, showing their respective conditions and targets. The CoT E2E model includes intermediate targets, highlighted in \textcolor{blue}{blue}. Here, $\underline{\langle\text{a\_tk}\rangle}$, $\underline{\langle\text{t\_tk}\rangle}$, and $\underline{\langle\text{s\_tk}\rangle}$ denote the audio, text, and speaker tokenizer identifiers, respectively.}
 \vskip -0.1in
\centering
\renewcommand{\arraystretch}{1.4}
\resizebox{\linewidth}{!}{
\begin{tabular}{lll}
\hline
\textbf{Task} & \textbf{Condition} & \textbf{Target} \\ \hline
\textbf{ASR} & $\underline{C}^{\text{asr}} = (\underline{\langle\text{asr}\rangle}, \underline{\langle\text{a\_tk}\rangle}, \underline{X}, \underline{\langle\text{t\_tk}\rangle})$ & $\underline{S}^{\text{asr}}$ \\[0.3em]
\textbf{T2T} & $\underline{C}^{\text{t2t}} = (\underline{\langle\text{t2t}\rangle}, \underline{\langle\text{t\_tk}\rangle}, \underline{\hat{S}}^{\text{asr}}, \underline{\langle\text{t\_tk}\rangle)}$ & $\underline{S}^{\text{res}}$ \\[0.3em]
\textbf{TTS} & $\underline{C}^{\text{tts}} = (\underline{\langle\text{tts}\rangle}, \underline{\langle\text{t\_tk}\rangle}, \underline{\hat{S}}^{\text{res}}, \underline{\langle\text{s\_tk}\rangle}, \underline{X}^{\text{spk}}, \underline{\langle\text{a\_tk}\rangle})$ & $\underline{Y}$ \\[0.3em]
\textbf{E2E} & $\underline{\langle\text{s2s}\rangle}, \underline{\langle\text{a\_tk}\rangle}, \underline{X}, \underline{\langle\text{s\_tk}\rangle}, \underline{X}^{\text{spk}}, \underline{\langle\text{a\_tk}\rangle}$ & $\underline{Y}$ \\[0.3em]\midrule
\textbf{CoT E2E} & $\underline{C}^{\text{asr}}, \textcolor{blue}{\underline{S}^{\text{asr}}}, \underline{C}^{\text{t2t}}, \textcolor{blue}{\underline{S}^{\text{res}}}, \underline{C}^{\text{tts}}$ & $\underline{Y}$ \\ \hline
\end{tabular}
}
\vskip -0.2in
\label{tab:task_templates}
\end{table}
\begin{table*}[h]
    \centering
    \caption{Semantic Quality on Switchboard Eval 2000 and Fisher: Our CoT-based E2E model produces more coherent responses (p $<$ 0.05)  than conventional E2E systems and matches the performance of the state-of-the-art LLM, SmolLM.}
    \vskip -0.1in
    \resizebox{0.9\linewidth}{!}{
    \begin{tabular}{l|ccc|ccc}
        \hline
        \textbf{Model} & \multicolumn{3}{c|}{SWBD} & \multicolumn{3}{c}{Fisher}\\
        & \textbf{ROUGE-1/2/L (↑)} & \textbf{METEOR (↑)} & \textbf{Perplexity (↓)} & 
          \textbf{ROUGE-1/2/L (↑)} & \textbf{METEOR (↑)} & \textbf{Perplexity (↓)} \\
        \hline
        SmolLM v2-1.7B (GT Transcript) & 14.0 / \textbf{2.3} / \textbf{11.5} & 12.9 & \hphantom{0}25.4 & 15.3 / \textbf{3.4} / \textbf{12.5} & 13.9 & \hphantom{0}\textbf{17.9}\\
        SpeechLM E2E & 10.5 / 0.8 / \hphantom{0}8.4 & \hphantom{0}9.7 & 302.2 & 11.3 / 1.3 / \hphantom{0}8.9 & 10.1 & 138.7\\\midrule
        SpeechLM CoT E2E &\\
        \hphantom{0}Text Response ($\hat{S}^{\text{res}}$) & 14.2 / 1.3 / 10.2 & 14.2 & \hphantom{0}37.2  & \textbf{17.6} / 3.2 / 12.2 & 17.6 & \hphantom{0}33.0 \\
         \hphantom{00} w/o Post-process (\textit{ablation}) & 12.3 / 1.3 / \hphantom{0}8.6 & 14.8 & \hphantom{0}24.1 & 14.6 / 2.9 / 10.0 & 17.7 & \hphantom{0}16.1 \\
         \hphantom{000} using GT Transcript $\hat{S}^{\text{asr}}$ (\textit{ablation}) & 12.3 / 1.4 / \hphantom{0}8.8 & 15.0 & \hphantom{0}24.5 & 15.0 / 3.0 / 10.2 & 17.8 & \hphantom{0}16.6 \\
         \hphantom{0} Speech Response ($\hat{Y}$) & 14.2 / 1.3 / 10.3 & 12.9 & \hphantom{0}28.5 & \textbf{17.6} / 3.2 / 12.3 & \textbf{17.7} & \hphantom{0}18.5\\
        SWBD+Fisher train $\hat{Y}$ (\S~\ref{subsec:dataset}) & \textbf{14.6} / 1.5 / 10.4 & \textbf{14.9} & \hphantom{0}\textbf{22.2} & \xmark & \xmark & \xmark\\
        \hline
    \end{tabular}
    }
    \vskip -0.2in
    \label{tab:text_response_results}
\end{table*}
\vspace{-3pt}
\subsection{Text and Audio Joint Tokenizer}
\vspace{-3pt}
\label{sec:joint_tokenizer}
Our CoT training procedure utilizes autoregressive generation of a joint text/audio token, as defined in~\cite{jinchuan2024speechlm}.
First, speech feature sequence $X$, text $S$, and special token $\langle \text{token} \rangle$ are tokenized $(\texttt{Tok}(\cdot))$ to the following discrete representation:
\begin{align}
    \underline{X} = \texttt{Tok}(X), \ 
    \underline{S} = \texttt{Tok}(S), \ 
    \underline{\langle \text{token} \rangle} = \texttt{Tok}(\langle \text{token} \rangle),
\end{align}
where the underline $\underline{\quad}$ indicates the tokenized value.
$\underline{X}$ $=$ $(\underline{\mathbf{x}}_1, \underline{\mathbf{x}}_2, ..., )$ 
is a sequence of $M$-dimensional discrete audio codec and its $t$-th vector is represented as
$\underline{\mathbf{x}}_t = [\underline{x}_t^1, ..., \underline{x}_t^M]^\top$.
$\underline{x}_t^1$ is reserved for a speech semantic token~\cite{audiolm}.
Similarly, $\underline{S}$ is composed of $j$-th discrete vector $\underline{\mathbf{s}}_j = [\underline{s}_j, \emptyset, ..., \emptyset]^\top$, where $\emptyset$ pads null values from the 2nd to $M$th dimension, as text tokens are one-dimensional.
$\langle \text{token} \rangle$ is also converted similarly to text tokens.
All tokens belong to a \textit{shared discrete vector space}, i.e., $\underline{\mathbf{x}}_t, \underline{\mathbf{s}}_j, \underline{\langle \text{token} \rangle} \in \mathcal{V}^M$, where $\mathcal{V}^M$ comprises the union of text, speech semantic and acoustic token vocabularies, along with special tokens.
This formulation enables speech, text, and special tokens to be autoregressively predicted using an LM.

\vspace{-3pt}
\subsection{Training Sequence Format}
\vspace{-3pt}
All the training and decoding steps of our model, introduced in Section~\ref{sec:method}, are tokenized into a single sequence and uniformly predicted.
Tab.~\ref{tab:task_templates} presents our sequence formats for each subtask in the cascaded pipeline (ASR, text response generation, and TTS) and for E2E and CoT E2E spoken dialogue training.
Task tokens are denoted as $\underline{\langle\text{asr}\rangle}$, $\underline{\langle\text{t2t}\rangle}$, $\underline{\langle\text{tts}\rangle}$, and $\underline{\langle\text{s2s}\rangle}$, representing ASR, text response generation, TTS, and E2E spoken dialogue tasks, respectively.
Notably, the CoT E2E model does not introduce new task tokens and consists of \textcolor{blue}{intermediate targets} that represent outputs from the first and second decoding stages (Sec.~\ref{sec:method}). Prompts for each decoding stage are carefully designed to align closely with SpeechLM’s pre-training objectives: ASR ($\underline{C}^{\text{asr}}$), text LM ($\underline{C}^{\text{t2t}}$), and TTS ($\underline{C}^{\text{tts}}$) as shown in Tab.~\ref{tab:task_templates}, enabling better training efficiency.\footnote{Other prompt designs resulted in sub-optimal performance.}

\vspace{-3pt}
\subsection{Training and Inference}
\vspace{-3pt}
\label{subsec:training_and_inference}
During CoT-post training, we compute the loss only on the ``target'' sequences: including intermediate targets $\textcolor{blue}{\underline{S}^{\text{asr}}}$, $\textcolor{blue}{\underline{S}^{\text{res}}}$ and  
final target $\underline{Y}$.  
During inference, our decoding process consists of three steps: \circled{1} The SpeechLM generates the ASR transcript $\textcolor{blue}{\underline{\hat{S}}^{\text{asr}}}=\argmax P(\underline{S}^{\text{asr}} | \underline{C}^{\text{asr}})$. 
\circled{2} The generated ASR transcript is then incorporated to generate text response, sampled as $\textcolor{blue}{\hat{\underline{S}}^{\text{res}}} \sim P(\underline{S}^{\text{res}} | \underline{C}^{\text{asr}},\hat{\underline{S}}^{\text{asr}}, \underline{C}^{\text{t2t}})$ using top-k sampling.
\circled{3} The predicted text response is similarly used to construct prompt (Tab.~\ref{tab:task_templates}) for sampling the final speech response $\hat{\underline{Y}}\sim P(\underline{Y} | \underline{C}^{\text{asr}},\hat{\underline{S}}^{\text{asr}}, \underline{C}^{\text{t2t}}, \hat{\underline{S}}^{\text{res}}, \underline{C}^{\text{tts}})$. By employing joint tokenization (Sec.~\ref{sec:joint_tokenizer}), our approach performs CoT-based inference (Eq.~\ref{eq:final_formulation}) within a standard LLM inference framework.

\section{Experiments}
\label{sec:experiment}
\vspace{-3pt}
\subsection{Datasets}
\vspace{-3pt}
\label{subsec:dataset}
For our experiments, we focus exclusively on \emph{real} human-human conversation datasets, selecting 2 widely used corpora: Switchboard~\cite{Switchboard} ($\approx$ 300 hours) and Fisher~\cite{cieri2004fisher} ($\approx$ 2000 hours). 
For Switchboard, we use the Eval2000 dataset for evaluation, while for Fisher, we follow Dialog GSLM~\cite{Dialog_GSLM}, splitting the dataset into 98:1:1 for train, dev, and evaluation, respectively. Additionally, we train the model in a \emph{combined} setting (SWBD+Fisher train) using the \textbf{entire} Fisher dataset and the Switchboard training set and evaluate it on the Eval2000 dataset.
Before training, we apply 3 preprocessing steps: \circled{1} We merge silence-separated utterances within a single speaker's turn to preserve conversational coherence. \circled{2} We remove very short utterances (fewer than five words) to improve linguistic context and response quality. \circled{3} We truncate all utterances to a maximum length of 30 seconds.

We evaluate spoken dialogue systems across semantic quality and audio quality and intelligibility of response. For semantic quality, we report ROUGE~\cite{lin2004rouge} and METEOR~\cite{banerjee2005meteor} scores, using human references as ground truth, alongside perplexity~\cite{jelinek1977perplexity}, computed using GPT-2~\cite{gpt-2}. 
For E2E models, semantic quality is evaluated by transcribing synthesized speech $\hat{Y}$ using Whisper large ~\cite{whisper}. For our CoT model, we also report metrics on the intermediate text response $\hat{S}^{\text{res}}$ (Eq.~\ref{eq:final_formulation}).
We utilize the VERSA toolkit~\cite{shi2024versa}, measuring intelligibility similarly through Whisper hypotheses and evaluating audio quality using UTMOS~\cite{saeki22c_interspeech}. We evaluate conversation-level performance by extracting emotion vectors from both synthesized outputs and ground-truth responses using Emo2Vec~\cite{ma2023emotion2vec}, then measure their cosine similarity. We rank all spoken dialogue systems based on their emotional alignment and compute the average rank (``Emotion Rank'') across all utterances. Finally, we report the model sizes for both cascaded and E2E dialogue systems.
\begin{table}[t]
    \centering
    \caption{Audio Quality and Intelligibility: Our CoT E2E model, trained on Fisher and Switchboard, matches single-speaker TTS (LJSpeech VITS) performance with a high-quality speaker prompt (``Spk prompt''). ``GT'' indicates decoding with ground-truth responses $\underline{S}^{\text{res}}$ and (abl) denote ablation study.}
    \vskip -0.1in
    \resizebox{\linewidth}{!}{
    \begin{tabular}{l|cc|cc}
        \hline
        \textbf{Model} & \multicolumn{2}{c|}{SWBD} & \multicolumn{2}{c}{Fisher}\\
        & \textbf{WER (↓)} & \textbf{UTMOS (↑)} & \textbf{WER (↓)} & \textbf{UTMOS (↑)} \\
        \hline
        LJSpeech VITS (GT) & \hphantom{0}8.5 & 4.19 & \hphantom{0}9.5 & 4.14\\
        SpeechLM & & \\
        \hphantom{0}Pre-train (GT) & 47.8 & 2.08 & 55.7 & 1.93 \\
        \hphantom{0}Finetune (GT) & 30.5 & 2.21 & 32.6 &  2.14\\
        \hphantom{0}E2E  & \xmark & 2.03  & \xmark & 2.02  \\\midrule
        \hphantom{0}CoT E2E  & 13.6 & 3.55 & 12.5 & 3.32\\
        \hphantom{00}w/o Spk prompt (\textit{abl}) & 15.1 & 2.05 & 15.5 & 2.04\\
        \hphantom{000}w/ GT (\textit{abl}) & 15.5 & 2.25 & 13.2 & 2.23 \\
        \hphantom{0000}w/o Post-process (\textit{abl}) & 42.0 & 2.10 & 36.5 & 2.10\\        
        \hphantom{0}SWBD+Fisher train (\S~\ref{subsec:dataset}) & \hphantom{0}9.1 & 3.40 & \xmark & \xmark\\
        \hline
    \end{tabular}
    }
    \vskip -0.2in
    \label{tab:tts_results}
\end{table}
\vspace{-3pt}
\subsection{Baseline and Experimental Setups}
\vspace{-3pt}
\label{subsec:experiment_setup}
We compare our CoT E2E system on response quality against strong task-specific baselines\footnote{Baselines are not fine-tuned on spoken dialogue datasets.}: SmolLM v2 1.7B-Instruct~\cite{allal2025smollm2smolgoesbig} for text generation and the single-speaker TTS model (\textbf{LJSpeech VITS})~\cite{hayashi2020espnet} from ESPnet-TTS.
Our system applies the CoT-based post-training strategy to a pre-trained open-source SpeechLM~\cite{jinchuan2024speechlm}. For TTS, we also evaluate the pre-trained SpeechLM in two settings: zero-shot (\textbf{SpeechLM Pre-train}) and fine-tuned (\textbf{SpeechLM Fine-tune}).\footnote{SpeechLM is not instruction fine-tuned on conversational data, so we exclude it from the text response evaluation.}
We further compare with cascaded and E2E spoken dialogue systems using conversation-level metrics. The cascaded systems combine: SpeechLM (ASR), SmolLM v2 (text response), and SpeechLM (TTS), evaluated in both zero-shot (\textbf{SpeechLM Cascaded (Pre-train)}) and fine-tuned (\textbf{SpeechLM Cascaded (Fine-tune)}) modes.
We also compare with a traditional E2E system (\textbf{SpeechLM E2E}, Sec.~\ref{sec: problem formulation}), where SpeechLM is trained end-to-end to directly predict speech outputs.

Our models are implemented in PyTorch, with all experiments conducted using the ESPnet~\cite{espnet,ESPnet-SLU} toolkit. 
The pre-trained SpeechLM leverages the SmolLM2 1.7B text LLM for initialization. 
We adopt the delay interleave architecture~\cite{musicgen} for multi-stream language modeling.
For audio tokenization, we concatenate codec and SSL tokens frame-by-frame. Specifically, we utilize ESPnet-Codec~\cite{shi2024espnet}\footnote{\url{https://huggingface.co/ftshijt/espnet\_codec\_dac\_large\_v1.4\_360epoch}} for codec tokenization and XEUS\footnote{\url{https://huggingface.co/espnet/xeus}, K-means tokenizer trained on the last-layer representation with 5k clusters}~\cite{chen2024towards} for SSL tokenization.
For decoding (Sec.~\ref{subsec:training_and_inference}), ASR uses greedy search followed by post-processing to remove hallucinations, while text response generation employs top-k sampling (k=30, temperature = 0.8), followed by post-processing to remove hallucinations and constrain response length. For speech response generation, we apply top-k sampling (same as text response), and further post-process the outputs, computing top-10 samples and selecting the speech with the highest intelligibility to generated text response $\hat{S}^{\text{res}}$ (Eq.~\ref{eq:final_formulation}). We conducted Wilcoxon signed-rank and Signed Paired Comparison tests for statistical significance.
Models are trained using 4 NVIDIA H200 GPUs. 
We will release data processing, training and inference details as part of ESPnet~\cite{espnet,arora-etal-2025-espnet} toolkit.

\begin{table}[t]
    \centering
    \caption{Conversation Level Statistics on Switchboard: our CoT E2E model better captures emotion and is parameter efficient. }
    \vskip -0.1in
    \resizebox{\linewidth}{!}{
    \begin{tabular}{lcc}
        \hline
        \textbf{Model} & \textbf{Emotion Rank (↓)}  & \textbf{Model Size (↓)} \\
        \hline
        SpeechLM &  \\
        \hphantom{00}Cascaded (Pre-train) & 3.08  & 3.4B\\
        \hphantom{00}Cascaded (Fine-tune) & 2.57  & 5.1B\\\midrule
        \hphantom{00} E2E  & 2.59 & \textbf{1.7B}  \\
        \hphantom{00} CoT E2E (SWBD + Fisher)  & \textbf{1.77} & \textbf{1.7B} \\
        \hline
    \end{tabular}
    }
    \label{tab:conversation_results}
    \vskip -0.2in
\end{table}
\vspace{-10pt}
\section{Results and Discussion}
\textbf{Semantic Coherence and Audio Quality Results}:
\label{subsec:semantic_audio_result}
Tab.~\ref{tab:text_response_results} presents the semantic quality of responses. Even state-of-the-art LLM SmolLM2 struggles to achieve high ROUGE and METEOR scores (Tab.~\ref{tab:text_response_results}), reflecting the spontaneity and unpredictability of human-human conversations. Despite this, our CoT E2E model generates semantically coherent responses, performing on par with SmolLM2, even when SmolLM2 uses ground-truth transcripts. Whisper hypotheses produced from final speech response $\hat{Y}$ exhibit similar semantic quality to the intermediate text response $\hat{S}^{\text{res}}$.\footnote{While our post-processing increases perplexity, it prevents very long responses.} We observe no significant performance drop when using generated ASR transcripts $\underline{\hat{S}}^{\text{asr}}$ in place of ground-truth transcripts ($\underline{S}^{\text{asr}}$ in CoT E2E prompt (Tab.~\ref{tab:task_templates})).
Conventional E2E models (``SpeechLM E2E'') show poor overlap with human references and generate semantically incoherent sentences, as shown by their high perplexity scores. 

Next, we evaluated audio quality performance in Tab.~\ref{tab:tts_results}. The pre-trained SpeechLM performs significantly worse in both intelligibility and speech quality against a single-speaker VITS model, likely due to poor generalization on disfluent, conversational text and the low-quality audio in Switchboard, as the model uses speaker prompts $\underline{X}^{\text{spk}}$ (TTS prompt in Tab.~\ref{tab:task_templates}) from the corresponding speaker. Fine-tuning on conversational datasets improves both speech quality and intelligibility.
Our CoT post-training achieves similar performance improvements, further boosting TTS quality through the post-processing step (Sec.~\ref{subsec:experiment_setup}, ``Post-process'' in Tab.~\ref{tab:tts_results}). Additionally, replacing Switchboard speaker prompts with high-quality prompts from Librispeech (UTMOS Score $\approx$ 4.5) during inference (``Spk prompt'' in Tab.~\ref{tab:tts_results}) results in substantial gains in both intelligibility and audio quality.
For the E2E SpeechLM, we cannot compute intelligibility due to the absence of ground-truth text references. However, we observe that its synthesized audio quality is inferior to that of the CoT-based E2E model, further underscoring the advantages of multi-stage reasoning.
Finally, our CoT-based E2E model, trained on a combined dataset of Switchboard and Fisher, significantly  (p $<$ 0.05) outperforms all baselines and matches performance of the single-speaker VITS model,  with added flexibility for multi-speaker speech.
\begin{table}[t]
    \centering
    \caption{Ablation Study on intermediate task ASR: While ASR performance slightly decreases with CoT training, the quality of text responses improves (Tab.~\ref{tab:text_response_results}).}
    \vskip -0.1in
    \resizebox{\linewidth}{!}{
    \begin{tabular}{l|cc|cc}
        \hline
        \textbf{Model} & \multicolumn{2}{c|}{SWBD} & \multicolumn{2}{c}{Fisher}\\
        & \textbf{WER (↓)} & \textbf{CER (↓)} & \textbf{WER (↓)} & \textbf{CER (↓)} \\
        \hline
        OWSM CTC (3.2) & 17.2 & 12.7 &12.9 & \hphantom{0}8.6 \\
        SpeechLM & & \\
        \hphantom{00} Pre-train & 18.3 & 13.3 & 14.3 & \hphantom{0}9.6\\
        \hphantom{00} Finetune & 17.8 & 13.1  & 13.4 & \hphantom{0}9.0\\ \midrule
        \hphantom{00} CoT E2E ($\hat{S}^{\text{asr}}$) & 17.6 & 13.1 & 19.5 & 14.1 \\
        \hphantom{0000} SWBD + Fisher Train & 22.2 & 17.0 & \xmark & \xmark\\
        \hline
    \end{tabular}
    }
    \vskip -0.2in
    \label{tab:asr_results}
\end{table}

\textbf{Conversation Level Analysis}:Tab.~\ref{tab:conversation_results} shows a conversation-level analysis of our CoT-based E2E model, comparing its performance with various cascaded and E2E spoken dialogue systems. Our analysis reveals that CoT modeling enhances dialogue expressiveness, as shown by higher emotion similarity with ground-truth human responses. Additionally, our CoT model demonstrates strong parameter efficiency while delivering performance comparable to the modules in cascaded systems (Tab.~\ref{tab:text_response_results},~\ref{tab:tts_results}), showcasing its applicability to on-device scenarios.
While its latency remains similar to cascaded systems due to multi-stage decoding, future work will focus on reducing latency through techniques like quantization.

\textbf{Ablation Study: Intermediate ASR task}:
Tab.~\ref{tab:asr_results} report the ASR performance of our CoT spoken dialogue model, comparing it with the OWSM 3.1~\cite{owsm} and pre-trained SpeechLM. The pre-trained SpeechLM performs competitively with OWSM, with fine-tuning further reducing WER. While CoT post-training slightly impacts ASR performance due to hallucinations, note that ASR serves only as an intermediate task. Importantly, Tab.~\ref{tab:text_response_results} demonstrates that the response quality remains robust against ASR hallucinations.

\vspace{-5pt}
\section{Conclusion}
\vspace{-5pt}
We propose a CoT-based formulation for E2E spoken dialogue systems, achieving competitive performance with task-specific baselines, while producing more coherent responses with superior audio quality than conventional E2E models. Our CoT-based approach also surpasses cascaded systems in parameter efficiency and emotional expressiveness. Future work will explore methods to enable real-time interaction and support ``speaking while listening''~\cite{kyutai2024moshi,xie2024miniomnilanguagemodelshear}.

\vspace{-3pt}
\section{Acknowledgement}
\vspace{-3pt}
Experiments of this work used the Bridges2 system at PSC and Delta system at NCSA through allocations CIS210014 and IRI120008P from the Advanced Cyberinfrastructure Coordination Ecosystem: Services \& Support (ACCESS) program, supported by National Science Foundation grants \#2138259,\#:2138286, \#:2138307, \#:2137603, and \#:2138296.

\newpage

\section{References}
{
\printbibliography
}
\end{document}